\documentclass[11pt,a4paper]{article}

\usepackage{acl}

\usepackage{times}
\usepackage[T1]{fontenc}
\usepackage[utf8]{inputenc}

\usepackage{graphicx}   
\usepackage{booktabs}   
\usepackage{multirow}
\usepackage{multicol}
\usepackage{amsmath,amssymb}
\usepackage{hyperref}
\usepackage{rotating} 

\usepackage{adjustbox}

\title{IndicTriMix: Developing Language Identification Datasets and Models for Tri-Language Code-Mixing}

\author{Pruthwik Mishra\textsuperscript{1}, Rudra Trivedi\textsuperscript{1}, Avi Patel\textsuperscript{1}, Ashok Urlana\textsuperscript{2}, Shrikant Malviya\textsuperscript{1} \\
  Sardar Vallabhbhai National Institute of Technology, Surat, India\textsuperscript{1} \\
  TCS Research, Hyderabad, India\textsuperscript{2}\\
  \texttt{\{u24ai068, u24ai071, pruthwikmishra\}@aid.svnit.ac.in} \\
  \texttt{shrikant@coed.svnit.ac.in,ashok.urlana@tcs.com}}
\date{June 2026}

\begin{document}
\maketitle

\begin{abstract}
Language identification in code-mixed  text, largely observed in social media, is highly essential when users frequently switch between multiple languages within a single utterance. Accurately identifying the languages of code-mixed tokens becomes an urgent necessity. Traditional language identification models, designed for monolingual text, are not well suited for token-level language identification in code-mixed settings. We formulate the task as a sequence labeling problem and fine-tune contextual transformer-based models MuRIL and XLM-RoBERTa best suited for Indian languages. We evaluate these systems on three different data configurations (Hindi, Gujarati, and Bengali) to predict language labels for individual tokens. We release a benchmark for language identification in code-mixed tokens with manually annotated test sets. We propose two approaches of code-mixed generation using parallel sentences of three languages.
The trained models demonstrate the effectiveness of contextual embeddings for token-level language identification in multilingual social media text. For reproducibility and to facilitate future research, we publicly release our fine-tuned models\footnote{\url{https://huggingface.co/avi-patel/indictrimix-models}}, datasets\footnote{\url{https://huggingface.co/datasets/avi-patel/indictrimix-dataset}}, and source code\footnote{\url{https://github.com/Avi4306/IndicTriMix}}.

\end{abstract}

\section{Introduction}
Social media platforms have seen a massive surge in multilingual user engagement, where code-mixing is the alternating use of two or more languages within a single conversation or utterance. For multilingual societies like India, around 7\% of the population speaks three languages~\cite{trilingual}. 
Individuals frequently intermix English with regional languages such as Hindi, Bengali, and Gujarati.

While conventional language identification (LID) tools perform reliably at the document or sentence level for monolingual texts, they fail significantly when applied to code-mixed user-generated content. Code-mixed text demands a token-level fine-grained classification framework, transforming the task into a structured sequence labeling problem.

Code-mixing, in principle, is a phenomenon that can blend two or more languages or dialects in a single utterance. But the token-level language identification task has been limited to two languages~\cite{amin2023marathienglishcodemixedtextgeneration,patra2018sentiment,barman2014code,bali2014borrowing,sheth2026beyond,kodali2022symcom} with a single matrix language and one embedded language. Very few works~\cite{goswami-etal-2023-offmix,raihan-etal-2023-sentmix,raihan2024emomix} have explored three-language code-mixed data, focusing on downstream tasks such as offensive language identification, sentiment analysis, and emotion detection, respectively, and are limited to English, Hindi, and Bengali. We attempt to develop language-agnostic techniques that can be applied and generalized to any three languages given 3-way parallel corpora.

In this paper, we present our models for language tag detection in multi-lingual code-mixed settings. Building upon the Transformer architecture~\cite{NIPS2017_3f5ee243}, we fine-tune multilingual transformer models, specifically MuRIL~\cite{khanuja2021murilmultilingualrepresentationsindian} and XLM-RoBERTa~\cite{conneau-etal-2020-unsupervised}, under three training configurations and evaluate each model on the corresponding in-domain development and test sets, framing the task as a token classification problem. Our models are trained to process code-mixed social media sentences and assign accurate linguistic tags to every constituent token.

\section{Related Work}

\subsection{Traditional Approaches to Language Identification}
Token-level language identification has traditionally used dictionary-based methods, 
n-gram language models, and sequence models such as Conditional Random Fields~\cite{645530.655813}, Maximum Entropy Markov Models~\cite{ratnaparkhi-1996-maximum}, Structured Perceptrons~\cite{collins-2002-discriminative}, and Hidden Markov Models~\cite{brants-2000-tnt}. However, code-mixed social media text introduces additional challenges due to transliteration, spelling variations, and informal language~\cite{barman2014code}.

\subsection{Neural and Transformer-based Approaches}
Neural approaches such as BiLSTM~\cite{GRAVES2005602} models, BiLSTM-CRF models~\cite{huang2015bidirectionallstmcrfmodelssequence}, and BiLSTM-CNN-CRFs~\cite{ma-hovy-2016-end} improved contextual language identification for code-mixed text~\cite{chaitanya2018word,mandal-singh-2018-language}, while recent studies have shown the effectiveness of transformer-based models for multilingual and code-mixed language identification~\cite{thara2021transformer,deka2023deep}.

\subsection{Multilingual Pretrained Models for Code-Mixed Text}
Multilingual pretrained models such as MuRIL~\cite{khanuja2021murilmultilingualrepresentationsindian} and XLM-RoBERTa~\cite{conneau-etal-2020-unsupervised} provide strong contextual representations for multilingual NLP and are particularly relevant to Indian code-mixed text~\cite{indiccmix}. In this work, we compare MuRIL and XLM-RoBERTa under combined multilingual, ENG-HIN-BEN, and ENG-HIN-GUJ training configurations.

\subsection{Research Gap}
To the best of our knowledge, no existing work has addressed code-mixed language identification involving three languages simultaneously, with prior studies largely restricted to bilingual code-mixed settings.



\section{Dataset Description}
We create two kinds of datasets: one rule-based  and another LLM-based that involves human annotations by experts. We limit the scope of this task to two types of trilingual code-mixing: ENG-HIN-GUJ and ENG-HIN-BEN. We utilize ISO-639-2, or three-lettered, language tags to represent the languages as shown in Table~\ref{tab:lang_id}. An additional tag ``UNI'' denotes the symbols and punctuations appearing in the corpus. These language tags act as labels for the language identification task. We sample sentences from the IndicCMix~\cite{indiccmix} dataset and label individual tokens using both techniques, as mentioned above. IndicCMix consists of 1.1 million sentences, where each of the unique English sentences (104,809) is translated into 11 Indic languages. We select this corpus as the base for code-mixing in three languages because the sentences in Indic languages are code-mixed in nature. Additionally, it provides the text in roman, which eliminates the need for transliteration. This substantially reduces the risk of error propagation caused by external transliteration tools.
\begin{table}[ht]
    \centering
    \begin{tabular}{l|l}
       \textbf{Identifier}  & \textbf{Language}\\
       \hline
       BEN& Bengali\\
       ENG& English\\
       GUJ & Gujarati\\
       HIN & Hindi\\
       UNI&Punctuation\\\hline
    \end{tabular}
    \caption{Identifier to Language Mapping}
    \label{tab:lang_id}
\end{table}
Three distinct data configurations are used to evaluate model performance over various linguistic compositions. The data statistics are shown in Table~\ref{tab:rule_based} and Table~\ref{tab:llm_gen}.
\subsection{Rule-Based Approach}
We use high-precision rules to identify the languages in the code-mixed sentences. To create a robust language identification model, a dataset that covers all varieties of code-mixing is required. The dataset must also contain sentences without any kind of code-mixing to enable the model to detect the language of monolingual text. We sample monolingual corpora from high-quality, publicly available corpora for our task. We choose Hindi~\cite{bhat2017hindi} and Bengali\footnote{\url{https://ltrc.iiit.ac.in/showfile.php?filename=downloads/kolhi/}} \cite{tandonsharma2017unity} corpora from publicly released dependency treebanks and Gujarati~\cite{bhattacharjee2025corilenrichingindianlanguage} from publicly released parallel corpora. IndicCMix dataset is chosen for code-mixing of two and three languages. Each Indic sentence (in our case, Hindi, Gujarati, and Bengali) consists of its romanized form, its native form, and the original source English sentence. We inspect English words in each Indic sentence, and if they are also found in the corresponding English source sentence, they are labeled as ``ENG''. Other tokens in the sentence are labeled based on the language of the sentence. The symbols and punctuations are tagged as ``UNIV''. For code-mixing involving three languages, we utilize the 3-way parallel corpora involving either English, Hindi, and Gujarati or English, Hindi, and Bengali available in IndicCMix data. We combine phrases from three languages. To avoid generating sentences with a single dominant matrix language and the other two languages having only one or two words, we have enforced constraints on the minimum number of tokens in each language. As the languages of each sentence is already known, this rule-based technique ensures that the languages of words in the code-mixed sentence are unambiguously tagged. The dataset details using this approach are presented in Table~\ref{tab:rule_based}.
\begin{table}[ht]
\centering
\begin{tabular}{l|l|l} \toprule
\textbf{Code-Mixing} & \textbf{\#Train} & \textbf{\#Dev}\\\midrule
ENG-HIN-BEN& 9142 & 796\\
ENG-HIN-GUJ & 9129 & 795\\ \bottomrule
\end{tabular}
\caption{Data Statistics Using Rule-Based Code-Mixing in terms of Sentences}
\label{tab:rule_based}
\end{table}

\begin{table}[ht]
\centering
\begin{tabular}{l|l|l} \toprule
\textbf{Code-Mixing} & \textbf{\#Dev} & \textbf{\#Test}\\\midrule
ENG-HIN-BEN&484  &550 \\
ENG-HIN-GUJ & 482 & 550\\ \bottomrule
\end{tabular}
\caption{Data Statistics Using LLM Generated Code-Mixing With Human Annotation in terms of Sentences}
\label{tab:llm_gen}
\end{table}

\begin{table}[htb]
\centering
\small
\begin{tabular}{p{0.95\linewidth}}
\toprule
\textbf{Instruction:} You are an expert linguist specializing in code-mixing (intra-sentential language switching). I will provide you with parallel sentences in English, Hindi, Gujarati, and Bengali, along with a starting index number. Your task is to generate two distinct, contextually meaningful code-mixed sentences for each sample provided. \\[4pt]

\textit{Code-Mixing Requirements:} \\
\textbf{EN-HI-GU:} Create a code-mixed sentence seamlessly blending English, Hindi, and Gujarati vocabulary and grammar. \\
\textbf{EN-HI-BN:} Create a code-mixed sentence seamlessly blending English, Hindi, and Bengali vocabulary and grammar. \\[4pt]

\textit{Linguistic Constraints (Crucial):} \\
\textbf{Vary the Sequence:} Do NOT follow a rigid sequence (e.g., always starting with English, then Hindi, then the regional language). \\
\textbf{Natural Flow:} Shuffle the syntax and vary where the English, Hindi, and regional language phrases appear in the structure. Mimic how real-life multilingual speakers naturally weave languages together depending on the context and focus of the sentence. \\
\textbf{Meaningful:} Ensure that despite the variations and blending, the final sentences are grammatically coherent and contextually meaningful. \\[4pt]

\textit{Output Formatting:} \\
Output the result for each sample as a single line containing a Python-style dictionary. Include an index key that increments sequentially starting from the provided starting index. Do not include row numbers, labels, or any conversational text outside the dictionaries. \\
\bottomrule
\end{tabular}
\caption{Prompt used for generating three-way code-mixed (EN-HI-GU and EN-HI-BN) sentences via the language model.}
\label{tab:prompt_gemini1}
\end{table}

\subsection{LLM-Based Approach}
For generating sentences with trilingual code-mixing, we use \texttt{Gemini 2.5 Pro}, which is a proprietary model. Using this technique, around 500 sentences are generated in both the dev and test sets. Two language experts manually annotate the language tag of each token for all the generated sentences in both the code-mixed settings. Each of the expert is a trilingual speaker with at least a postgraduate level of education. The statistics are shown in Table~\ref{tab:llm_gen}. The prompt used for the code-mixed generation is detailed in the Table~\ref{tab:prompt_gemini1}.

\section{Assessing Quality of Generated Code-Mixed Sentences}
Parallel sentences act as the main pivot for our code-mixed generation approaches. In order to assess the quality of the generated code-mixed sentences, we utilize various pretrained models that represent sentences using multilingual shared embeddings. For our study, BertScore~\cite{Zhang2020BERTScore}, Sentence BERT~\cite{reimers-gurevych-2019-sentence}, LaBSE~\cite{feng-etal-2022-language} embeddings are used to measure the semantic similarity between each generated code-mixed sentence and its corresponding language specific sentence. In Sentence-BERT, specifically MPNET~\footnote{\url{https://huggingface.co/sentence-transformers/paraphrase-multilingual-mpnet-base-v2}} model and LaBSE, the semantic closeness is evaluated by computing the cosine similarity between vector representations of two sentences. BertScore, or Bert F1-Score evaluates the maximal similarities at a token level. The semantic similarities of the code-mixed sentences generated by both approaches with each of the methods exceed 0.85 on average. This indicates high fluency and faithfulness of the generated sentences. The details are added in Appendix~\ref{sec:sem_score} under Table~\ref{tab:sem_score}.
\section{Methodology}
We treat language tagging at the token level as a sequence labeling problem. Given a sentence that can be viewed as a sequence of tokens $S=(w_1,w_2,\ldots,w_n)$, we seek to produce an analogous sequence of labels $Y=(y_1,y_2,\ldots,y_n)$ with $y_i \in \{\text{BEN}, \text{ENG}, \text{GUJ}, \text{HIN},\text{UNI}\}$.

\subsection*{Step-by-Step Procedure}
The pipeline we have used for compilation, tokenization, and evaluation of our models has been set up in the following manner:
\begin{enumerate}
    \item \textbf{Data Parsing}: Text data formatted in a CoNLL-style format is read in iteratively. The sentences are dynamically collected, and each sequence is separated using either an empty line or a separation boundary.
    \item \textbf{Tokenization and Subword Alignment}: Tokenization was performed using the Hugging Face Transformers package~\cite{wolf-etal-2020-transformers} with the pretrained tokenizer corresponding to each model. Since a single word may be split into multiple subword tokens, while the annotations are provided at the word level, word-to-subword alignment was required. 
    \item \textbf{Label Masking}: When a word was split into multiple subword tokens, its ground-truth label was assigned only to the first subword. All subsequent subwords belonging to the same word were assigned the ignore index \texttt{-100}, preventing them from contributing to the training loss. 
    \item \textbf{Data Batching and Dynamic Padding}: Extracted features are converted to structured dictionary mappings using HuggingFace framework. This ensures sequences within a batch are dynamically padded to match the longest element, optimizing compute times.
    \item \textbf{Supervised Fine-Tuning}: The downstream system feeds contextual hidden vectors from the transformer body into a linear token classification layer tasked with estimating cross-entropy distributions across the target tag configurations.
\end{enumerate}


\section{Model Architectures}
We evaluate two prominent multilingual transformer architectures for token-level language identification:

\subsection{MuRIL}
MuRIL (Multilingual Representations for Indian Languages)~\cite{khanuja2021murilmultilingualrepresentationsindian} is a pre-trained language model specifically designed to capture linguistic nuances across Indian languages and their code-mixed variations. Its architecture is particularly suited for handling the morphological and phonetic complexities of Indo-Aryan languages written in both native scripts and Latin transliteration.

\subsection{XLM-RoBERTa}
XLM-RoBERTa-base~\cite{conneau-etal-2020-unsupervised} is a cross-lingual transformer model trained on 100+ languages. It provides universal multilingual representations and serves as a strong baseline for comparison across diverse language pairs.

\section{Experimental Setup}
The models were implemented using PyTorch~\cite{NEURIPS2019_bdbca288} and the Hugging Face Transformers library~\cite{wolf-etal-2020-transformers}. Training employed mixed-precision (fp16) optimization to improve computational efficiency. The foundational settings used for our experiments are detailed in Table~\ref{tab:hyperparameters}.

\begin{table}[ht]
\centering
\small
\begin{tabular}{l|l}
\toprule
\textbf{Hyperparameter} & \textbf{Value} \\
\midrule
Base Architectures & MuRIL, XLM-RoBERTa-Base \\
Learning Rate & $2 \times 10^{-5}$ \\
Batch Size (Train/Eval) & 16 \\
Total Training Epochs & 10 \\
Weight Decay & 0.01 \\
Max Sequence Length & 128 tokens \\
Optimization Metric & Macro F\_1 score \\
\bottomrule
\end{tabular}
\caption{Hyperparameter settings for fine-tuning.}
\label{tab:hyperparameters}
\end{table}

\begin{table*}[ht]
\centering
\small
\setlength{\tabcolsep}{4pt}
\renewcommand{\arraystretch}{1.3}
\begin{tabular}{lcccccccccc}
\toprule
& \multicolumn{5}{c}{\textbf{MuRIL}}
& \multicolumn{5}{c}{\textbf{XLM-RoBERTa}} \\
\cmidrule(lr){2-6}\cmidrule(lr){7-11}

& \multicolumn{1}{c}{\textbf{BEN}}
& \multicolumn{1}{c}{\textbf{ENG}}
& \multicolumn{1}{c}{\textbf{GUJ}}
& \multicolumn{1}{c}{\textbf{HIN}}
& \multicolumn{1}{c}{\textbf{UNI}}
& \multicolumn{1}{c}{\textbf{BEN}}
& \multicolumn{1}{c}{\textbf{ENG}}
& \multicolumn{1}{c}{\textbf{GUJ}}
& \multicolumn{1}{c}{\textbf{HIN}}
& \multicolumn{1}{c}{\textbf{UNI}} \\
\midrule

\textbf{ENG-HIN-BEN-DEV-RB}
& 0.984
& \textbf{0.987}
& -
& 0.992
& 0.999
& \textbf{0.984}
& 0.987
& -
& \textbf{0.993}
& \textbf{1.0} \\

\textbf{ENG-HIN-GUJ-DEV-RB}
& -
& \textbf{0.99}
& \textbf{0.985}
& \textbf{0.991}
& 0.996
& -
& 0.989
& 0.981
& 0.988
& \textbf{0.998} \\

\textbf{ENG-HIN-BEN-DEV-LLM}
& \textbf{0.981}
& \textbf{0.99}
& -
& 0.408
& -
& 0.97
& 0.984
& -
& \textbf{0.423}
& \textbf{0.997} \\

\textbf{ENG-HIN-BEN-TEST-LLM}
& \textbf{0.975}
& \textbf{0.993}
& -
& \textbf{0.773}
& \textbf{1.0}
& 0.968
& 0.99
& -
& 0.771
& \textbf{1.0} \\

\textbf{ENG-HIN-GUJ-DEV-LLM}
& -
& \textbf{0.991}
& 0.913
& 0.547
& 0.996
& -
& 0.991
& \textbf{0.921}
& \textbf{0.617}
& \textbf{0.997} \\

\textbf{ENG-HIN-GUJ-TEST-LLM}
& -
& \textbf{0.996}
& 0.923
& 0.757
& \textbf{1.0}
& -
& 0.993
& \textbf{0.924}
& \textbf{0.773}
& \textbf{1.0} \\

\bottomrule
\end{tabular}
\caption{Results from Combined training on development and test sets. Values represent the $F_1$-score for each language class.}
\label{tab:combined_results}
\end{table*}
\subsection{Training Procedure and Convergence}
We trained both MuRIL and XLM-RoBERTa on three separate data configurations:
\begin{enumerate}
    \item \textbf{Combined Data}: Merged multilingual data from all language pairs.
    \item \textbf{ENG-HIN-BEN Configuration}: Data containing English, Hindi, and Bengali code-mixed text.
    \item \textbf{ENG-HIN-GUJ Configuration}: Data containing English, Hindi, and Gujarati code-mixed text.
\end{enumerate}

All models were fine-tuned for 10 training epochs using the configuration described in Table~\ref{tab:hyperparameters}. The rule-based development (Dev RB) set was used as the validation set, and model evaluation was performed at the end of each epoch. The checkpoint achieving the best validation macro $F_1$-score was retained for subsequent evaluation.

\begin{table*}[ht]
\centering
\small
\setlength{\tabcolsep}{4pt}
\renewcommand{\arraystretch}{1.3}
\begin{tabular}{lcccccccc}
\toprule
& \multicolumn{4}{c}{\textbf{MuRIL}}
& \multicolumn{4}{c}{\textbf{XLM-RoBERTa}} \\
\cmidrule(lr){2-5}\cmidrule(lr){6-9}

& \multicolumn{1}{c}{\textbf{BEN}}
& \multicolumn{1}{c}{\textbf{ENG}}
& \multicolumn{1}{c}{\textbf{HIN}}
& \multicolumn{1}{c}{\textbf{UNI}}
& \multicolumn{1}{c}{\textbf{BEN}}
& \multicolumn{1}{c}{\textbf{ENG}}
& \multicolumn{1}{c}{\textbf{HIN}}
& \multicolumn{1}{c}{\textbf{UNI}} \\
\midrule

\textbf{ENG-HIN-BEN-DEV-RB}
& 0.983
& 0.985
& \textbf{0.992}
& 0.999
& \textbf{0.983}
& \textbf{0.986}
& 0.991
& \textbf{1.0} \\

\textbf{ENG-HIN-BEN-DEV-LLM}
& \textbf{0.979}
& \textbf{0.991}
& 0.339
& 0.996
& 0.977
& 0.989
& \textbf{0.364}
& \textbf{1.0} \\

\textbf{ENG-HIN-BEN-TEST-LLM}
& \textbf{0.974}
& \textbf{0.992}
& \textbf{0.764}
& \textbf{1.0}
& 0.97
& 0.991
& 0.762
& \textbf{1.0} \\

\bottomrule
\end{tabular}
\caption{Results from ENG-HIN-BEN training on development and test sets. Values represent the $F_1$-score for each language class.}
\label{tab:enghinben_results}
\end{table*}
\section{Results and Evaluation}
Our fine-tuned systems achieved robust performance on the evaluation sets. We compare MuRIL and XLM-RoBERTa under three training configurations:
(i) combined multilingual training,
(ii) ENG-HIN-BEN-specific training, and
(iii) ENG-HIN-GUJ-specific training.
Each model is evaluated on the corresponding development and test sets. The following sections present detailed evaluation results for each configuration.

\subsection{Model Results}

\subsubsection{Combined Data Training}

The token-level classification results for MuRIL and XLM-RoBERTa trained on the combined multilingual dataset and evaluated on the rule-based (RB) and LLM-generated development and test sets are presented in Table~\ref{tab:combined_results}.

\subsubsection{ENG-HIN-BEN Training}

The token-level classification results for MuRIL and XLM-RoBERTa trained specifically on the ENG-HIN-BEN multilingual dataset and evaluated on the rule-based (RB) and LLM-generated development and test sets are presented in Table~\ref{tab:enghinben_results}.

\subsubsection{ENG-HIN-GUJ Training}

The token-level classification results for MuRIL and XLM-RoBERTa trained specifically on the ENG-HIN-GUJ multilingual dataset and evaluated on the rule-based (RB) and LLM-generated development and test sets are presented in Table~\ref{tab:enghinguj_results}.

\begin{table*}[ht]
\centering
\small
\setlength{\tabcolsep}{4pt}
\renewcommand{\arraystretch}{1.3}
\begin{tabular}{lcccccccc}
\toprule
& \multicolumn{4}{c}{\textbf{MuRIL}}
& \multicolumn{4}{c}{\textbf{XLM-RoBERTa}} \\
\cmidrule(lr){2-5}\cmidrule(lr){6-9}

& \multicolumn{1}{c}{\textbf{ENG}}
& \multicolumn{1}{c}{\textbf{GUJ}}
& \multicolumn{1}{c}{\textbf{HIN}}
& \multicolumn{1}{c}{\textbf{UNI}}
& \multicolumn{1}{c}{\textbf{ENG}}
& \multicolumn{1}{c}{\textbf{GUJ}}
& \multicolumn{1}{c}{\textbf{HIN}}
& \multicolumn{1}{c}{\textbf{UNI}} \\
\midrule

\textbf{ENG-HIN-GUJ-DEV-RB}
& \textbf{0.988}
& \textbf{0.983}
& \textbf{0.989}
& 0.996
& 0.987
& 0.981
& 0.988
& \textbf{0.997} \\

\textbf{ENG-HIN-GUJ-DEV-LLM}
& \textbf{0.994}
& 0.923
& 0.603
& 0.996
& 0.992
& \textbf{0.927}
& \textbf{0.645}
& \textbf{0.998} \\

\textbf{ENG-HIN-GUJ-TEST-LLM}
& \textbf{0.996}
& \textbf{0.928}
& 0.778
& \textbf{1.0}
& 0.993
& 0.927
& \textbf{0.779}
& \textbf{1.0} \\

\bottomrule
\end{tabular}
\caption{Results from ENG-HIN-GUJ training on development and test sets. Values represent the $F_1$-score for each language class.}
\label{tab:enghinguj_results}
\end{table*}

\subsection{Discussion}
We observe several key patterns across the three training configurations and two architectures:
\begin{enumerate}
    \item \textbf{Language-Pair-Specific Training}: Models trained on a specific language pair achieve consistently strong performance on the corresponding evaluation set, indicating that specialized training effectively captures language-specific characteristics of the code-mixed data.

    \item \textbf{Combined Data Generalization}: Training on the combined multilingual dataset produces competitive performance across both language pairs, demonstrating that a single model can effectively learn shared multilingual representations while maintaining strong overall performance.

    \item \textbf{Architecture Comparison}: MuRIL generally achieves slightly better performance than XLM-RoBERTa across the evaluated configurations, particularly under combined-data training. This advantage is consistent with MuRIL's pre-training emphasis on Indian languages and its suitability for multilingual and code-mixed text involving Indian languages. However, the performance difference varies across language configurations and evaluation splits, with XLM-RoBERTa achieving comparable performance in several cases.

    \item \textbf{Language-Specific Challenges}: The difficulty of language identification varies across evaluation settings. While the rule-based development sets achieve consistently high F\_1-scores, the LLM-generated sets show greater variation, particularly for HIN. This may be attributed to transliteration, lexical overlap, and limited class support in some splits, where a small number of errors can substantially affect the F\_1-score. In contrast, UNI is consistently recognized with near-perfect F\_1-scores.
\end{enumerate}
\section{Conclusion}
In this study, we present token classification frameworks for language tag detection using both MuRIL and XLM-RoBERTa transformer models trained on three distinct data configurations: combined multilingual data, ENG-HIN-BEN code-mixed text, and ENG-HIN-GUJ code-mixed text. Evaluation on the development and test sets enabled us to compare the effectiveness of language-pair-specific models with a single model trained on combined multilingual data. The experimental results demonstrate that contextual word embeddings provide a robust foundation for managing sequence boundaries and structural changes in multi-lingual code-mixed sentences. 
\section{Limitations}
\subsection*{Data Deficiencies and Target Label Imbalances}
A primary challenge identified during model evaluation is the effect of uneven label distributions across different evaluation subsets. When an evaluation slice contains a highly imbalanced distribution of language instances, particularly when a language has very few supporting examples, the model's $F_1$-score can become sensitive to precision and recall variations. This can make performance estimates less stable for languages with limited representation in a particular evaluation set.

\subsection*{Phonetic Interference and Lexical Errors in Transliteration}
The use of textual abbreviations, slang words, and formatting anomalies characteristic of informal code-mixed text makes language classification more challenging. Furthermore, several Indo-Aryan languages exhibit similar phonetic characteristics when represented using the Latin alphabet (Romanization), which can blur the boundaries between language classes. This transliteration-related ambiguity may affect the overall performance of the system for certain language instances. Future work could address this issue by incorporating additional lexical or phonetic information.

\subsection*{Subword-Level Label Alignment}
Since the language-identification annotations are provided at the word level, words that are segmented into multiple subword tokens require an alignment strategy. In our implementation, the original word-level label is assigned only to the first subword, while subsequent subwords are assigned the ignore index \texttt{-100}. Although this avoids assigning the same word-level label multiple times and prevents trailing subwords from contributing directly to the training loss, it also means that these subword representations do not receive direct supervision from the corresponding word-level label. More sophisticated word-to-subword labeling strategies could be explored in future work.


\bibliography{custom}
\appendix

\begin{table*}[htb]
\centering
\resizebox{\textwidth}{!}{%
\begin{tabular}{|l|cccc|cccc|cccc|}
\hline
\multirow{2}{*}{} 
& \multicolumn{4}{c|}{\textbf{BertScore-F1}}
& \multicolumn{4}{c|}{\textbf{LaBSE}}
& \multicolumn{4}{c|}{\textbf{MPNET}} \\ \cline{2-13}

& \textbf{ENG} & \textbf{HIN} & \textbf{BEN} & \textbf{GUJ}
& \textbf{ENG} & \textbf{HIN} & \textbf{BEN} & \textbf{GUJ}
& \textbf{ENG} & \textbf{HIN} & \textbf{BEN} & \textbf{GUJ} \\ \hline

\textbf{ENG-HIN-BEN-dev-RB}
& 0.915 & 0.936 & 0.919 & -
& 0.80 & 0.885 & 0.861 & -
& 0.773 & 0.877 & 0.854 & - \\ \hline

\textbf{ENG-HIN-GUJ-dev-RB}
& 0.906 & 0.935 & - & 0.929
& 0.79 & 0.896 & - & 0.9
& 0.753 & 0.883 & - & 0.889 \\ \hline

\textbf{ENG-HIN-BEN-dev-LLM}
& 0.911 & 0.9 & 0.925 & -
& 0.859 & 0.792 & 0.825 & -
& 0.837 & 0.77 & 0.839 & - \\ \hline

\textbf{ENG-HIN-BEN-test-LLM}
& 0.911 & 0.894 & 0.918 & -
& 0.837 & 0.776 & 0.831 & -
& 0.845 & 0.782 & 0.839 & - \\ \hline

\textbf{ENG-HIN-GUJ-dev-LLM}
& 0.913 & 0.905 & - & 0.908
& 0.855 & 0.815 & - & 0.824
& 0.831 & 0.798 & - & 0.818 \\ \hline

\textbf{ENG-HIN-GUJ-test-LLM}
& 0.913 & 0.908 & - & 0.902
& 0.841 & 0.817 & - & 0.811
& 0.830 & 0.815 & - & 0.821 \\ \hline

\end{tabular}%
}
\caption{Semantic Similarity Scores of Code-Mixed Sentences with Language-Wise Parallel Sentences. RB $\rightarrow$ Rule-Based Approach and LLM $\rightarrow$ LLM Approach.}
\label{tab:sem_score}
\end{table*}

\section{Appendix}
\subsection{Computation of Semantic Similarity Scores}
\label{sec:sem_score}
We compute the semantic similarity scores of the code-mixed sentences with English sentences and the romanized versions of Hindi, Gujarati, and Bengali sentences. From Table~\ref{tab:sem_score}, we can observe that the generated code-mixed sentences are both faithful and fluent with their language-specific counterparts evident from high scores.

\end{document}